# FedScore: A Privacy-Preserving Framework for Federated Scoring System Development

Siqi Li[1], Yilin Ning[1], Marcus Eng Hock Ong[2,3,4], Bibhas Chakraborty[1,2,5,6], Chuan Hong[6], Feng Xie[1,2], Han Yuan[1], Mingxuan Liu[1], Daniel M. Buckland[7], Yong Chen[8], Nan Liu[1,2,9]*

[1] Centre for Quantitative Medicine, Duke-NUS Medical School, Singapore, Singapore

[2] Programme in Health Services and Systems Research, Duke-NUS Medical School, Singapore, Singapore

[3] Health Services Research Centre, Singapore Health Services, Singapore, Singapore

[4] Department of Emergency Medicine, Singapore General Hospital, Singapore, Singapore

[5] Department of Statistics and Data Science, National University of Singapore, Singapore, Singapore

[6] Department of Biostatistics and Bioinformatics, Duke University, Durham, NC, USA

[7] Department of Emergency Medicine, Duke University School of Medicine, Durham, NC, USA

[8] Department of Biostatistics, Epidemiology and Informatics, University of Pennsylvania, Philadelphia, PA, USA

[9] Institute of Data Science, National University of Singapore, Singapore, Singapore

* Correspondence: Nan Liu, Centre for Quantitative Medicine, Duke-NUS Medical School, 8 College Road, Singapore 169857, Singapore. Phone: +65 6601 6503. Email: liu.nan@duke-nus.edu.sg

## Abstract

**Objective**

We propose FedScore, a privacy-preserving federated learning framework for scoring system generation across multiple sites to facilitate cross-institutional collaborations.

**Materials and Methods**

The FedScore framework includes five modules: federated variable ranking, federated variable transformation, federated score derivation, federated model selection and federated model evaluation. To illustrate usage and assess FedScore's performance, we built a hypothetical global scoring system for mortality prediction within 30 days after a visit to an emergency department using 10 simulated sites divided from a tertiary hospital in Singapore. We employed a pre-existing score generator to construct 10 local scoring systems independently at each site and we also developed a scoring system using centralized data for comparison.

**Results**

We compared the acquired FedScore model's performance with that of other scoring models using the receiver operating characteristic (ROC) analysis. The FedScore model achieved an average area under the curve (AUC) value of 0.763 across all sites, with a standard deviation (SD) of 0.020. We also calculated the average AUC values and SDs for each local model, and the FedScore model showed promising accuracy and stability with a high average AUC value which was closest to the one of the pooled model and SD which was lower than that of most local models.

**Conclusion**

This study demonstrates that FedScore is a privacy-preserving scoring system generator with potentially good generalizability.

# 1. Introduction

Cross-institutional collaboration has gained popularity in recent years as a way to accelerate medical research and facilitate quality improvement[1]. Widespread digitization efforts in the healthcare industry enable the use of data-driven evidence for clinical prediction models[2], which can be ideally built using centralized data pooled from as many sources as possible. Some examples of cross-regional collaborations include: the Collaborative European NeruroTrauma Effectiveness Research in Traumatic Brain Injury[3], the Genotype to Phenotype Databases[4], the Big Data in Cardiovascular Disease[5], the Ontario Prehospital Advanced Life Support[6], the Kaiser Permanente Research Bank[7] and the Pan-Asian Resuscitation Outcomes Study[8]. However, such partnerships require data sharing, which is typically laborious and time-consuming, and sometimes even impossible due to various privacy regulations[9], [10], for example the European Union General Data Protection Regulation[11].

Federated learning (FL), sometimes referred to as distributed learning or distributed algorithms, can avoid data sharing by collectively training algorithms without exchanging patient-level data[12], safeguarding patients' privacy by distributing the model-training to the data-owners and aggregating their results[13]. In addition to dismantling data silos, FL could also speed up the development of much-needed AI models[14]. For instance, during the COVID-19 pandemic, Dayan et al.[14] constructed a clinical outcomes prediction model across 20 institutes using FL. Luo et al. [15] studied the demographic and clinical factors that are associated with length of stay in COVID-19 patients using a lossless, one-shot FL algorithm[15]. There exist many applications of FL for medical image data, most of which use black box models from computer vision.

Interpretable models, on the contrary, have fewer instances of FL applications despite their popularity in clinical research.

As a type of interpretable risk scoring model[16], scoring systems have been employed in practically every diagnostic area of medicine[17], since they offer quick and simple risk assessments of numerous serious medical conditions without the use of a computer[16]. Some traditional scoring systems, such as the Glasgow Coma Scale[18] first described in 1974, rely heavily on clinician's domain expertise. More data-driven methods for building scoring systems have emerged in recent years, including the Supersparse Linear Integer Model[19] which can better deal with sparsity, approximal methods that are more computationally efficient[20], [21], and interfaces that enable flexible engagement of domain expertise, like the Interval Coded Scoring[22] and the AutoScore[23].

Regardless of development strategies, scoring systems have usually been created using single-source data, limiting application at other sites if the development data has insufficient sample size or is not representative. Although it is possible to develop scoring systems on pooled data[24], the process of doing such pooling, as noted previously, is time consuming and difficult to achieve due to privacy restrictions. As a result, frameworks for building scoring systems in a federated manner are needed to overcome such difficulties. To fill this gap, we propose FedScore, a first-of-its-kind framework for building federated scoring systems across multiple sites and demonstrated its efficacy and potential generalizability with a proof-of-concept experiment using real-world data.

## 2. Methods

Scoring systems are linear classification models that require users to add, subtract and multiply a few numbers in order to make a prediction[16], and have been widely utilized in the field of clinical decision-making[25]–[27] for risk stratification due to their interpretability and transparency. They can also assist in correcting physicians' misestimations of the probability of medical outcomes, which may be rather common[28]. Users frequently take into account a model's degree of parsimony when implementing clinical models[29], which means that a model is parsimonious if it is both sparse (i.e., it uses the least amount of variables possible) and has good prediction accuracy[29]. As an example, the AutoScore framework[23] is a computational tool to conveniently create such scores using machine learning methods, and has been well received by clinicians[30], [31], because it integrates domain knowledge with data driven evidence. However, regardless of the particulars of their generation of scoring systems and accounting for model interpretability, AutoScore and other similar methods only permit the development of scoring systems using one set of pooled data. To fully exploit the growing data sources and to create less biased models, we propose our FedScore framework to achieve good parsimony and interpretability for federated data, while complying with potential privacy restrictions.

### 2.1 FedScore Framework

The FedScore framework consists of five modules: 1) federated variable ranking; 2) federated variable transformation; 3) federated score derivation; 4) federated model selection and 5) federated model evaluation. The workflow of FedScore is illustrated in Figure 1.

1) Federated Variable Ranking

Variable selection is an essential step in the development of scoring systems for consideration of parsimony. To construct a global model across several sites, it is necessary to pre-identify a set of unified candidate variables. We employed random forests for variable importance measurement, which have been broadly utilized[32]–[36]. In FedScore, variable ranking is first performed independently via random forests at each local site, and then a global variable ranking is created by rearranging variables by their weighted ranks across all $K$ sites. Specifically, for a single variable $X_m$ where $1 \leq m \leq P$ and $P$ is the total number of predictors, let $q_j \in N$ denote its rank at site $j$, then its global ranking is obtained by mapping all values of $\sum_{j=1}^{K} w_j q_j$ for each site to the integer set $[1, P] \subset Z$. Here $w_j$ is the normalized weight for site $j$, satisfying $\sum_{j=1}^{K} w_j = 1$, and the default setting is $1/K$, indicating equal weights for all sites. Sample size based weight $w_i = S_j/S_0$ may also be applied, where $S_j$ is the sample size of site $j$ and $S_0$ is the total sample size. Users may also utilize other self-defined weights if they want to accommodate particular research aims.

2) Federated Variable Transformation

The creation of categorical variables allows for the modeling of nonlinear effects[16], [23], which has been widely applied[37]–[43] in the development of clinical scoring systems. Following this common practice, FedScore turns continuous variables into categorical variables after unified variable ranking is established. The maximum number of categories for such transformation is pre-determined (for example, choose 5 as a usual practice), and if the maximum is surpassed, categories are combined so that the requirement is met. In our study, the quantiles of continuous variables are set to be $0\%, k_1\%, k_2\%, k_3\%, k_4\%$, and $100\%$, where the default value of $k_1, k_2, k_3, k_4$ are 5, 20, 80 and 95. The unified cutoff for each continuous variable is calculated by weighting the $k$ values acquired at each site, in the same way that the global ranking is determined.

**3) Federated Score Derivation**

Binary outcomes are common in clinical decision making and logistic regression is a prominent method used for modelling such outcomes. Federated logistic regression may be achieved using different existing FL frameworks, and for demonstration purpose, we have employed a one-shot privacy preserving distributed algorithm called ODAL2[44], which is communication-efficient and has been demonstrated to have low bias and high statistical efficiency[44]. This algorithm utilizes information from the lead local site (data are accessible) with the first-order and second-order gradients of the likelihood function from remotes sites (data are not accessible) to construct an approximation of the global likelihood function. The global logistic regression coefficients can then be obtained by optimizing the approximate global likelihood function. Let $x_1, x_2, \dots x_{p-1}$ denote the $p-1$ predictors, $y$ denote a binary outcome, and the logistic regression model can be expressed as

$$logit(Pr(y = 1|x)) = x^T \beta$$

where $x = (1, x_1, x_2, \dots x_{p-1})^T$, β is the vector of intercept and slope coefficients, and $logit(t) = \log t/(1-t)$. Suppose a total of $N = \sum_{j=1}^{K} n_j$ identically and independently distributed (i.i.d.) observations are distributed at $K$ sites, then the log-likelihood function (LLR) of the global logistic regression by pooling data from all sites is

$$L(\beta) = \frac{1}{N} \sum_{j=1}^{K} \sum_{i=1}^{n_j} [Y_{ij} x_{ij}^T \beta - log\{1 + exp(x_{ij}^T \beta)\}]$$

The pooled estimator $\hat{\beta}$ can be obtained by optimizing $L(\beta)$. When data cannot be shared and the pooled likelihood function is not possible, approximation of the likelihood function is still achievable. The ODAL2 algorithm applies the idea of Taylor expansion, proposing to use first and second order gradient of LLR to perform the approximation[44]:

$$\tilde{L}^2(\beta) = L_1(\beta) + \{\nabla L(\bar{\beta}) - \nabla L_1(\bar{\beta})\}^T \beta + \frac{1}{2}(\beta - \bar{\beta})^T \{\nabla^2 L(\bar{\beta}) - \nabla^2 L_1(\bar{\beta})\}(\beta - \bar{\beta})$$

Here $\bar{\beta}$ is an initial value obtained from the regression model performed at the local site and stored for broadcasting to remote sites. $L_j(\beta) = \frac{1}{n_j} \sum_{i=1}^{n_j} [Y_{ij} x_{ij}^T \beta - log\{1 + exp(x_{ij}^T \beta)\}]$ is the LLR of the $j$-th site ($j = 1$ is assumed to the local site). $\nabla L(\bar{\beta}) = \sum_{j=1}^{K} n_j \nabla L_j(\bar{\beta})/N$ is the first gradient of the LLR $L(\bar{\beta})$ evaluated at $\bar{\beta}$ and $\nabla L_j(\bar{\beta}) = \frac{1}{n_j} \sum_{i=1}^{n_j} \{Y_{ij} - p_{ij}(\bar{\beta})\} x_{ij}$ is the first gradient of the LLR of site $j$, where $p_{ij}(\bar{\beta}) = 1 + exp(-x_{ij}^T \bar{\beta})^{-1}$ and $\nabla^2 L_j(\bar{\beta}) = \frac{1}{n_j} \sum_{i=1}^{n_j} p_{ij}(\bar{\beta})\{1 - p_{ij}(\bar{\beta})\} x_{ij} x_{ij}^T$ is the second gradient of LLR of site $j$. Both gradients are computed at each remote site and transferred back to the local site.

Finally, the global beta estimator of β is obtained by optimizing the surrogate likelihood function. This process for constructing the global model is one-shot[44] as illustrated in Figure 1, and neither of the shared files contain any patient level information, which guarantee privacy. Federated scores are obtained by having coefficients in the global logistic regression model rounded to integers and mapped to interval $[0, S_{max}]$, where $S_{max}$ is the maximum score pre-decided by users, e.g., 100.

**4) Federated Model Selection**

Model selection is performed using the parsimony plots generated on validation data. A general model selection criteria could be defined by maximizing $\Psi_m = \sum w_i \phi_i(p_1, p_2, p_3, \ldots p_m)$, where $w_i$ is the weight for site $i$ as described previously, $\phi_i$ measures a scoring's performance on the $i$th validation set (e.g. AUC value) and $m$ is a pre-specified number of total variables to include, which should be uniform across all sites. Different constraints can be added for the optimization task. For example, the total number of variables $m$ may not exceed an integer number $D$. The set of variables $\{p_1, p_2, \ldots p_m\}$ may also be set to satisfy certain subjective standard required by users. For instance, users may decide (based on domain knowledge) that a set of variables $\{x_1, x_2, \ldots x_q\}$, where $q \leq m$ must be included in the final scoring system regardless of the results provided by variable important analysis. Moreover, $\Psi$ may be maximized using a number of $d$ of variables that is smaller than $m$, as long as increasing the variable numbers from $d$ to $m$ has little impact on the change in $\Psi$: $|\Psi_m - \Psi_d| \leq \epsilon$, where the size of $\epsilon$ may be decided intuitively by users based on parsimony plots.

After final variables are confirmed based on the selected model, a new model is refitted via module 2) so that the final model is as parsimony as possible.

**5) Federated Model Evaluation**

The performance of the final model is validated on each site engaged in the FedScore framework. Following the $\Psi_m$ defined in step 4), the overall weighted performance of a federated score is $M_1 = \sum w_j \mu_j(p_1, p_2, p_3, \ldots p_m)$, where $\mu_j$ is the score's performance on $j$th testing set; and $M_2 = \left(\sum w_j (M_1 - \mu_j)^2\right)^{1/2}$ is a measurement of weighted performance variation across sites. A higher $M_1$ value and lower $M_2$ value indicate a score's better performance and generalizability.

The FedScore framework has been implemented in R 4.0.3 and code is available at https://github.com/nliulab/FedScore.

**2.2 Experiment**

The initial study cohort was formed by selecting emergency department (ED) visits in 2016 and 2017, using the EHR data of Singapore General Hospital (SGH) extracted from the SingHealth Electronic Health Intelligence System. A waiver of consent was granted for EHR data collection and the retrospective analysis, and the study has been approved by the Singapore Health Services' Centralized Institutional Review Board, with all data deidentified. After excluding patients under the age of 18 and those with missing values, the remaining cohort was randomly divided into 10 sites for demonstration purpose, in the proportion of 4%, 5%, 7%, 9%, 10%, 11%, 12%, 13%, 14%, and 15% respectively. Figure 2 depicts the process of cohort formation.

The outcome in this study was whether a patient died within 30 days after ED admissions. Candidate variables were determined based on a recent work[30], the study cohort of which was also obtained from SGH ED data. The candidate predictors include a total of 29 variables in 5 categories: (1) demographics information: age, sex and race; (2) PACS[45] triage categories (P1, P2, P3 and P4), shift time (8 AM to 4 PM, 4 PM to midnight, Midnight to 8 AM), and day of week (Friday, Monday, Weekend, Midweek); (3) vital signs: pulse (beats/min), respiration (times/min), peripheral capillary oxygen saturation ($SpO_2$; %), diastolic blood pressure (mm Hg), and systolic blood pressure (mm Hg); (4) comorbidities: myocardial infarction, congestive heart failure, peripheral vascular disease, stroke, dementia, chronic pulmonary disease, rheumatoid disease, peptic ulcer disease, diabetes, hemiplegia or paraplegia, kidney disease, and liver disease; (5) previous health care usage: ED visits in the past year, surgical procedures in the past year, ICU admissions in the past year, and high-dependency admissions in the past year.

AutoScore was used to create baseline models for local and pooled comparisons with our FedScore framework. Three groups of analysis were performed: 1) 10 local scores trained independently on each site; 2) one federated score trained using all sites without data sharing; 3) one pooled score generated using centralized data, which is the ideal case but usually impossible in most real world settings. All models were chosen based on corresponding parsimony plots, with a predefined criterion that the maximum number of variables in a model should not exceed 8 and adding more variables until there is no significant improvement in AUC. In order to perform straightforward comparisons, the cutoffs and weights used during scoring system development were default values specified in Section 2.1 and all processes involved were data-driven without refining that engaged expert knowledge from clinical practice.

## 3. Results

A total of 80,613 individual ED admission episodes were randomly divided into 10 sites, with sample size ranging from 3224 to 12,092 and the training, validation and testing sets of each site were obtained by randomly splitting at ratios of 70%, 10%, and 20% respectively, as shown in Figure 2. Table 1 summarizes the baseline characteristics of the overall and each site's cohort.

We compared the performance of the federated score developed by FedScore with the pooled score developed using all data and the 10 local scores independently developed at each site. Figure 3 depicts how each score performed on the testing datasets of each site, with twelve subplots. For each subplot, a scoring model's performance on each of the ten sites is presented in horizontal lines using the corresponding AUC values and its 95% confidence intervals (CIs). The vertical edges of the grey rectangular frame in each subplot reflect the mean of ten AUC values plus/minus their standard deviation (SD) and as a result, the width of each grey rectangular frame represents the degree of performance variation of a model across all sites. The detailed AUC values, CIs and SDs are reported in Supplementary eTable 1. The scoring tables for each model and corresponding parsimony plots were also provided in eTable 2 and eFigure 1 of the Supplementary Materials.

With the information presented in Figure 3 and eTable 1, we summarize the following main observations: 1) the federated score achieved good performance, with an average AUC value across all sites of 0.763, better than that of the local models and close to the one of the pooled model; 2) the AUC variance of federated score is among the smallest ones, and although the SDs for local models of site 2 and 5 appear to be slightly smaller, their averaged AUC values are lower; 3) the performance of the federated score on some sites are better than the model developed locally at that site (e.g. $0.7804 > 0.7300$ at site 7).

## 4. Discussion

FedScore is among the first frameworks that aims to generalize unified scores across multiple sites while preserving privacy. The scalable and adaptable architecture offers potential solutions for improving model generalizability and stability across isolated clinical datasets.

Whereas scoring systems have been widely utilized in clinical domains, few existing FL applications have focused on them despite their prevalence, reflecting the phenomenon that existing biomedical FL applications have a tendency to favor black box models[46] over more interpretable ML models. To meet physicians' expectations for model simplicity and transparency, FL applications of interpretable models require more customization and modification compared to black box model implementations with well-established FL frameworks available from the computer science community. A simple and straightforward scoring table with a lower AUC value for risk stratification, for example, would be preferred by clinicians over a black box model with a higher AUC value in the ED. As a result, more cautious designs for FL applications of interpretable models are required and FedScore deals with this issue by emphasizing model parsimony and enabling flexible process monitoring for users. Future FL applications in clinical sciences should take similar factors into account if the research questions favor transparent solutions rather than merely being concerned with model performance.

FL studies in the biomedical field differ from the ones in computer science, albeit sharing similar origins. In many standard engineering FL contexts, since a single client cannot create models independently, attention has been paid to technical details such as data partitioning schemes and various privacy mechanisms[47]. In clinical domains however, data are frequently formed at the hospital or institution level, making local models feasible in these cases. Under these circumstances, generalizability (models' ability to generalize their performance to a new setting[48]) and stability of global models relative to local models become more crucial, but these factors are not sufficiently considered in many existing FL frameworks that are being developed. The results in section 3 show that by a co-training process via FL, a global model prediction framework such as FedScore can achieve less variation than locally developed ones while still maintaining good performance. This benefit of FL is promising for medical research that seeks dependable high risk decision making.

Data constraints, such as biased data and small datasets are considered a source of ML misuse[49], yet investigating such misconduct is not as feasible as developing models. Despite the emphasis[48] on external validations, less than 10% of clinical prediction studies reported to have done so[50]. Instead of training a model on single site and subsequently testing and modifying it on other sites, constructing a model with sufficient and representative data through privacy-preserving means may be a more viable solution. FedScore and its future extensions could potentially aid in reducing model inconsistency across cohorts, leading to more trustworthy decision-making for medical research.

Although we have only used one binary outcome example for illustration, our FedScore framework is scalable and versatile, given that modules could be appropriately modified to accommodate different clinical research questions. For instance, the score derivation module could be modified to accommodate survival or ordinal outcomes, and additional privacy-preserving FL frameworks and topologies might also be added to offer more options. We anticipate that FedScore and its future extensions could together act as some foundations for creating more trustworthy clinical scoring systems in approaches that safeguard data privacy.

**Limitations and Future Work**

Results were obtained from a homogenous data splitted from a single source and without real-world heterogeneity across sites. FedScore may encounter problems with heterogeneous medical data because the current ODAL2 algorithm in module 3 requires that the data across different sites are homogeneous, similar to the majority of the FL and distributed methods currently in use[12], [51]. However, since the proposed framework is scalable, it can be continuously updated and enhanced by cutting-edge solutions that can better deal with the issue. Future work will involve international collaboration to develop the FedScore process with more heterogeneous datasets. We plan to extend the FedScore by incorporating the two state-of-art FL algorithms that account for the between-site heterogeneity. The first strategy is to use the dCLR algorithm[52], motivated from a novel pairwise conditional logistic regression, to estimate the common regression coefficients and then estimate the site-specific intercept locally for each site. The second strategy is to adopt the lossless, few-shot dPQL algorithm[53], which has been used to rank the performance of different hospitals while considering the case-mix situation across sites (i.e., different hospitals are treating different patients).

## 5. Conclusion

We have proposed FedScore, a privacy-preserving scoring systems and used a 30-day mortality prediction task to show proof-of-concept. We have demonstrated its potential to build effective federated clinical scores that are more generalizable, with lower performance variability across sites. FedScore is a first-of-its-kind framework for constructing scoring systems based on distributed algorithms, bridging a gap in current medical research. While demonstrated for binary outcomes, the application of FedScore can be extended for settings with other types of clinical outcomes and greater heterogeneity across sites with future developments in FL and clinical prediction methods, enabling its use in a wide range of different medical contexts.

**Table 1** Description of the study cohorts[a] (N = 80613)

| | All Sites | Site 1 | Site 2 | Site 3 | Site 4 | Site 5 | Site 6 | Site 7 | Site 8 | Site 9 | Site 10 |
|---|---|---|---|---|---|---|---|---|---|---|---|
| # Episodes | 80,613 | 3,224 | 4,031 | 5,643 | 7,255 | 8,061 | 8,867 | 9,674 | 10,480 | 11,286 | 12,092 |
| Age, mean (SD) | 63.51 (17.74) | 64.09 (17.35) | 63.54 (17.77) | 63.32 (17.78) | 63.80 (17.70) | 63.38 (17.70) | 63.24 (18.08) | 63.52 (17.72) | 63.59 (17.76) | 63.29 (17.67) | 63.65 (17.68) |
| Gender | | | | | | | | | | | |
| Female | 40492 (50.2) | 1599 (49.6) | 2015 (50.0) | 2830 (50.2) | 3604 (49.7) | 4034 (50.0) | 4528 (51.1) | 4899 (50.6) | 5258 (50.2) | 5705 (50.5) | 6020 (49.8) |
| Male | 40121 (49.8) | 1625 (50.4) | 2016 (50.0) | 2813 (49.8) | 3651 (50.3) | 4027 (50.0) | 4339 (48.9) | 4775 (49.4) | 5222 (49.8) | 5581 (49.5) | 6072 (50.2) |
| Race/ethnicity | | | | | | | | | | | |
| Chinese | 56966 (70.7) | 2289 (71.0) | 2847 (70.6) | 3956 (70.1) | 5125 (70.6) | 5763 (71.5) | 6273 (70.7) | 6769 (70.0) | 7358 (70.2) | 7941 (70.4) | 8645 (71.5) |
| Indian | 8888 (11.0) | 320 (9.9) | 452 (11.2) | 610 (10.8) | 811 (11.2) | 819 (10.2) | 1026 (11.6) | 1120 (11.6) | 1171 (11.2) | 1266 (11.2) | 1293 (10.7) |
| Malay | 9793 (12.1) | 386 (12.0) | 478 (11.9) | 711 (12.6) | 865 (11.9) | 1006 (12.5) | 1026 (11.6) | 1203 (12.4) | 1308 (12.5) | 1368 (12.1) | 1442 (11.9) |
| Others | 4966 (6.2) | 229 (7.1) | 254 (6.3) | 366 (6.5) | 454 (6.3) | 473 (5.9) | 542 (6.1) | 582 (6.0) | 643 (6.1) | 711 (6.3) | 712 (5.9) |
| PACS triage categories | | | | | | | | | | | |
| P1 | 19169 (23.8) | 778 (24.1) | 950 (23.6) | 1279 (22.7) | 1752 (24.1) | 1865 (23.1) | 2095 (23.6) | 2311 (23.9) | 2579 (24.6) | 2665 (23.6) | 2895 (23.9) |
| P2 | 44572 (55.3) | 1800 (55.8) | 2210 (54.8) | 3166 (56.1) | 3989 (55.0) | 4547 (56.4) | 4852 (54.7) | 5275 (54.5) | 5762 (55.0) | 6286 (55.7) | 6685 (55.3) |
| P3 and P4 | 16872 (20.9) | 646 (20.0) | 871 (21.6) | 1198 (21.2) | 1514 (20.9) | 1649 (20.5) | 1920 (21.7) | 2088 (21.6) | 2139 (20.4) | 2335 (20.7) | 2512 (20.8) |
| Shift time | | | | | | | | | | | |
| 8 AM to 4 PM | 42594 (52.8) | 1707 (52.9) | 2133 (52.9) | 3012 (53.4) | 3851 (53.1) | 4275 (53.0) | 4649 (52.4) | 5158 (53.3) | 5609 (53.5) | 5893 (52.2) | 6307 (52.2) |
| 4 PM to midnight | 28141 (34.9) | 1115 (34.6) | 1419 (35.2) | 1937 (34.3) | 2537 (35.0) | 2829 (35.1) | 3112 (35.1) | 3378 (34.9) | 3584 (34.2) | 3961 (35.1) | 4269 (35.3) |
| Midnight to 8 AM | 9878 (12.3) | 402 (12.5) | 479 (11.9) | 694 (12.3) | 867 (12.0) | 957 (11.9) | 1106 (12.5) | 1138 (11.8) | 1287 (12.3) | 1432 (12.7) | 1516 (12.5) |
| Day of week | | | | | | | | | | | |
| Friday | 11060 (13.7) | 445 (13.8) | 555 (13.8) | 763 (13.5) | 983 (13.5) | 1088 (13.5) | 1234 (13.9) | 1313 (13.6) | 1388 (13.2) | 1620 (14.4) | 1671 (13.8) |
| Midweek | 35464 (44.0) | 1435 (44.5) | 1781 (44.2) | 2590 (45.9) | 3178 (43.8) | 3614 (44.8) | 3880 (43.8) | 4272 (44.2) | 4588 (43.8) | 4820 (42.7) | 5306 (43.9) |
| Monday | 13311 (16.5) | 528 (16.4) | 640 (15.9) | 905 (16.0) | 1200 (16.5) | 1306 (16.2) | 1486 (16.8) | 1610 (16.6) | 1756 (16.8) | 1897 (16.8) | 1983 (16.4) |
| Weekend | 20778 (25.8) | 816 (25.3) | 1055 (26.2) | 1385 (24.5) | 1894 (26.1) | 2053 (25.5) | 2267 (25.6) | 2479 (25.6) | 2748 (26.2) | 2949 (26.1) | 3132 (25.9) |
| Vital Signs, mean (SD) | | | | | | | | | | | |
| Pulse, /min | 86.38 (18.41) | 86.14 (18.75) | 86.68 (18.26) | 85.89 (18.03) | 86.39 (18.58) | 86.26 (18.34) | 86.51 (18.59) | 86.49 (18.56) | 86.65 (18.38) | 86.36 (18.21) | 86.27 (18.42) |
| Respiration, /min | 18.27 (2.16) | 18.26 (2.16) | 18.31 (2.18) | 18.22 (2.05) | 18.28 (2.21) | 18.22 (2.11) | 18.28 (2.19) | 18.25 (2.14) | 18.30 (2.14) | 18.27 (2.18) | 18.27 (2.22) |
| $SpO_2$, % | 97.37 (4.11) | 97.40 (3.66) | 97.37 (4.42) | 97.25 (4.93) | 97.28 (4.50) | 97.34 (4.03) | 97.43 (3.83) | 97.40 (3.95) | 97.40 (3.85) | 97.42 (3.65) | 97.32 (4.45) |
| Blood pressure, mm Hg | | | | | | | | | | | |
| Diastolic | 72.58 (14.06) | 72.62 (14.07) | 72.32 (13.78) | 72.70 (13.71) | 72.78 (14.09) | 72.67 (14.14) | 72.63 (14.20) | 72.54 (14.25) | 72.62 (14.18) | 72.54 (14.07) | 72.44 (13.87) |
| Systolic | 137.46 (27.98) | 137.95 (28.34) | 136.91 (27.23) | 137.56 (27.78) | 137.64 (27.98) | 137.73 (28.14) | 137.12 (27.72) | 137.45 (28.36) | 137.79 (28.36) | 137.48 (27.93) | 137.13 (27.70) |
| Comorbidities | | | | | | | | | | | |
| Myocardial infarction | 5095 (6.3) | 207 (6.4) | 262 (6.5) | 335 (5.9) | 462 (6.4) | 512 (6.4) | 540 (6.1) | 621 (6.4) | 666 (6.4) | 699 (6.2) | 791 (6.5) |
| Congestive heart failure | 8841 (11.0) | 363 (11.3) | 453 (11.2) | 600 (10.6) | 767 (10.6) | 901 (11.2) | 956 (10.8) | 1044 (10.8) | 1212 (11.6) | 1214 (10.8) | 1331 (11.0) |
| Peripheral vascular disease | 4649 (5.8) | 181 (5.6) | 219 (5.4) | 298 (5.3) | 429 (5.9) | 479 (5.9) | 522 (5.9) | 528 (5.5) | 654 (6.2) | 631 (5.6) | 708 (5.9) |
| Stroke | 9527 (11.8) | 400 (12.4) | 498 (12.4) | 653 (11.6) | 838 (11.6) | 969 (12.0) | 1055 (11.9) | 1141 (11.8) | 1260 (12.0) | 1307 (11.6) | 1406 (11.6) |

| | | | | | | | | | | |
|---|---|---|---|---|---|---|---|---|---|---|
| Dementia | 2854 (3.5) | 133 (4.1) | 137 (3.4) | 195 (3.5) | 253 (3.5) | 278 (3.4) | 334 (3.8) | 322 (3.3) | 369 (3.5) | 386 (3.4) | 447 (3.7) |
| Chronic pulmonary disease | 7089 (8.8) | 290 (9.0) | 378 (9.4) | 491 (8.7) | 634 (8.7) | 678 (8.4) | 804 (9.1) | 841 (8.7) | 887 (8.5) | 994 (8.8) | 1092 (9.0) |
| Rheumatoid disease | 1165 (1.4) | 53 (1.6) | 54 (1.3) | 85 (1.5) | 97 (1.3) | 125 (1.6) | 126 (1.4) | 136 (1.4) | 155 (1.5) | 164 (1.5) | 170 (1.4) |
| Peptic ulcer disease | 2455 (3.0) | 106 (3.3) | 121 (3.0) | 167 (3.0) | 222 (3.1) | 263 (3.3) | 224 (2.5) | 300 (3.1) | 307 (2.9) | 359 (3.2) | 386 (3.2) |
| Diabetes | | | | | | | | | | | |
| None | 50165 (62.2) | 2004 (62.1) | 2526 (62.7) | 3601 (63.9) | 4491 (61.9) | 5007 (62.1) | 5529 (62.3) | 6018 (62.2) | 6448 (61.5) | 7020 (62.2) | 7521 (62.2) |
| Diabetes without chronic complications | 3640 (4.5) | 147 (4.6) | 189 (4.7) | 273 (4.8) | 334 (4.6) | 339 (4.2) | 414 (4.7) | 447 (4.6) | 502 (4.8) | 459 (4.1) | 536 (4.4) |
| Diabetes with complications | 26808 (33.3) | 1073 (33.3) | 1316 (32.6) | 1769 (31.3) | 2430 (33.5) | 2715 (33.7) | 2924 (33.0) | 3209 (33.2) | 3530 (33.7) | 3807 (33.7) | 4035 (33.4) |
| Hemiplegia or paraplegia | 3594 (4.5) | 156 (4.8) | 188 (4.7) | 269 (4.8) | 345 (4.8) | 373 (4.6) | 396 (4.5) | 411 (4.2) | 456 (4.4) | 445 (3.9) | 555 (4.6) |
| Kidney disease | 19142 (23.7) | 747 (23.2) | 947 (23.5) | 1312 (23.3) | 1753 (24.2) | 1923 (23.9) | 2080 (23.5) | 2264 (23.4) | 2534 (24.2) | 2700 (23.9) | 2882 (23.8) |
| Liver disease | | | | | | | | | | | |
| None | 75111 (93.2) | 3006 (93.2) | 3764 (93.4) | 5251 (93) | 6779 (93.5) | 7510 (93.1) | 8288 (93.4) | 8968 (92.7) | 9745 (93) | 10520 (93.2) | 11280 (93.3) |
| Mild liver disease | 3937 (4.9) | 151 (4.7) | 202 (5.0) | 276 (4.9) | 342 (4.7) | 392 (4.9) | 423 (4.8) | 490 (5.1) | 534 (5.1) | 538 (4.8) | 589 (4.9) |
| Severe liver disease | 1565 (1.9) | 67 (2.1) | 65 (1.6) | 116 (2.1) | 134 (1.8) | 159 (2.0) | 156 (1.8) | 216 (2.2) | 201 (1.9) | 228 (2.0) | 223 (1.8) |
| Health care use, mean (SD) | | | | | | | | | | | |
| Emergency admissions in the past year | 1.07 (2.40) | 1.03 (2.34) | 1.06 (2.38) | 1.04 (2.38) | 1.06 (2.26) | 1.07 (2.44) | 1.05 (2.35) | 1.09 (2.52) | 1.12 (2.51) | 1.04 (2.25) | 1.06 (2.46) |
| Operations in the past year | 0.29 (0.98) | 0.27 (0.88) | 0.28 (0.97) | 0.28 (1.04) | 0.29 (0.97) | 0.30 (1.00) | 0.29 (0.99) | 0.29 (0.98) | 0.30 (1.02) | 0.29 (0.96) | 0.28 (0.92) |
| ICU admissions in the past year | 0.03 (0.28) | 0.03 (0.31) | 0.02 (0.22) | 0.03 (0.37) | 0.03 (0.33) | 0.03 (0.32) | 0.02 (0.25) | 0.03 (0.29) | 0.02 (0.25) | 0.03 (0.26) | 0.03 (0.28) |
| HD admissions in the past year | 0.08 (0.44) | 0.08 (0.46) | 0.08 (0.45) | 0.08 (0.47) | 0.08 (0.44) | 0.07 (0.42) | 0.08 (0.46) | 0.08 (0.44) | 0.08 (0.43) | 0.08 (0.45) | 0.08 (0.45) |
| Outcome | | | | | | | | | | | |
| 30 day mortality | 4249 (5.3) | 158 (4.9) | 207 (5.1) | 294 (5.2) | 400 (5.5) | 405 (5.0) | 450 (5.1) | 478 (4.9) | 512 (4.9) | 640 (5.7) | 705 (5.8) |

Abbreviations:
HD, high-dependency; ICU, intensive care unit; PACS, Patient Acuity Category Scale; SpO$_2$, oxygen saturation as measured by pulse oximetry.
[a] Data are presented as count (percentage) of patients unless otherwise indicated.

**Figure 1**. Flowchart of the FedScore framework.

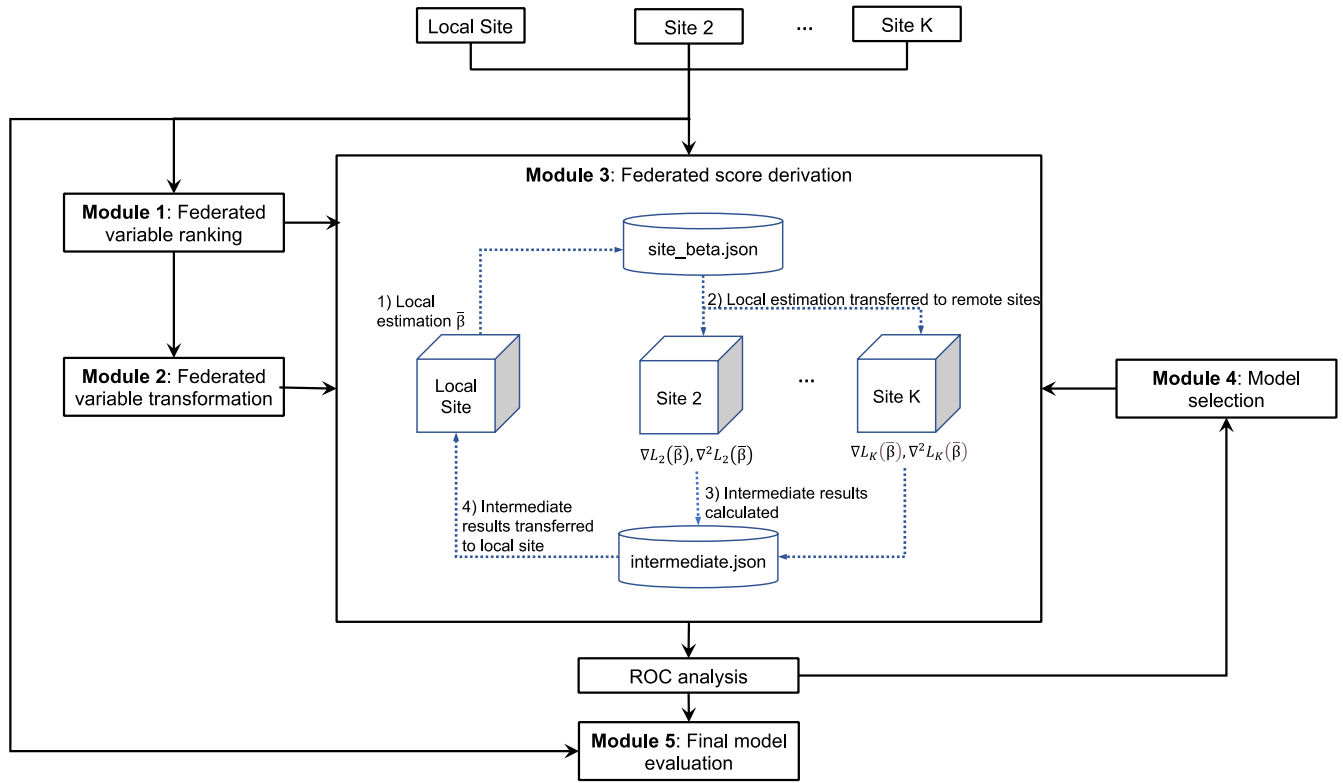

**Figure 2**. Flowchart of the study cohorts' formation. SGH: Singapore General Hospital

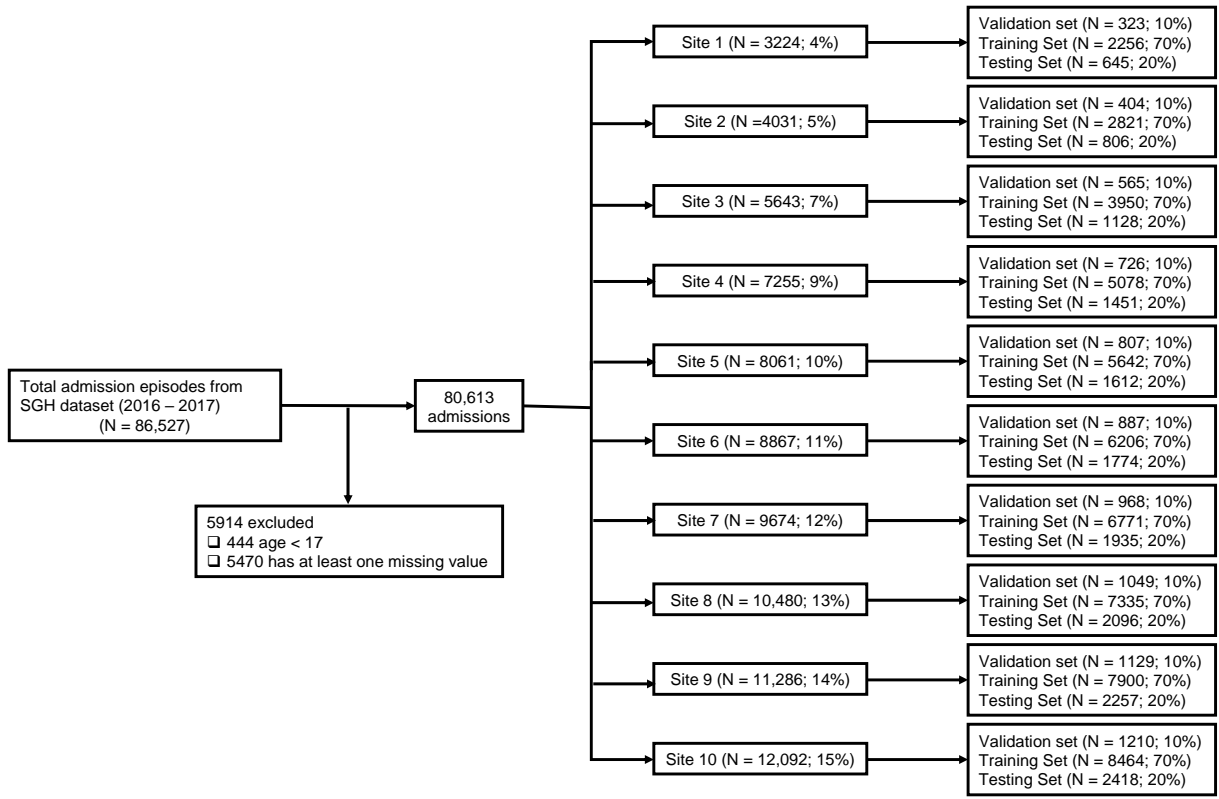

**Figure 3.** Comparison of performance of FedScore and local scores.

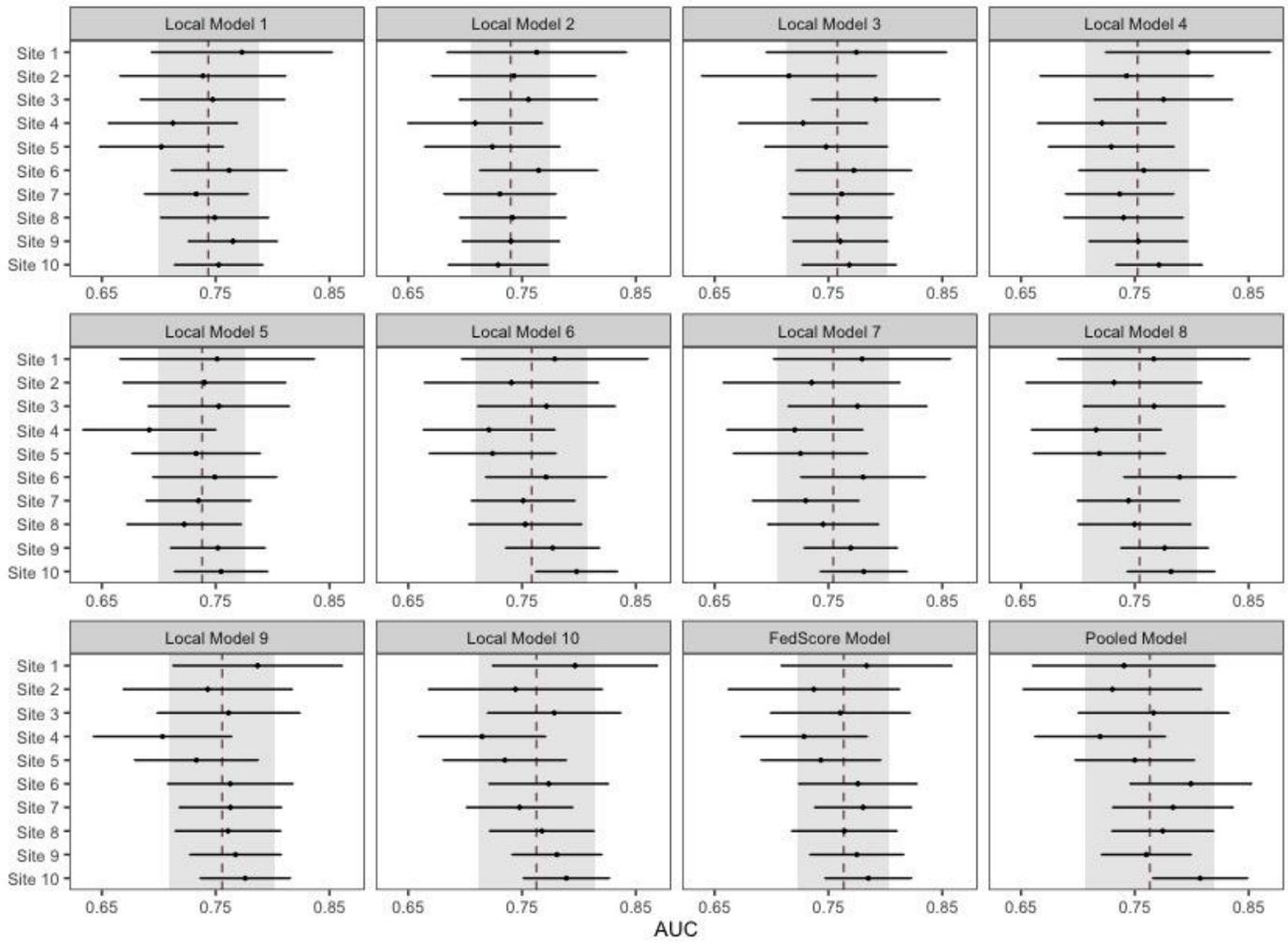

# Supplementary Materials

**eTable 1:** Comparison of performance of FedScore model with baseline models.

**eTable 2:** Scoring tables
    (a) - (j)**:** Scoring tables of local models generated on site 1 to site 10 independently via AutoScore.
(k): Scoring table of federated model generated via FedScore.
(l) : Scoring table of pooled model generated via AutoScore

**eFigure 1:** Parsimony plots
(a) - (j)**:** Parsimony plots of local models generated on site 1 to site 10 independently via AutoScore.
(k): Parsimony plot of federated model generated via FedScore.
(l): Parsimony plot of pooled model generated via AutoScore

**eTable 1** Comparison of performance of FedScore model with baseline models.

| Model | Number of variables | Testing Data | | | | | | | | | | | | | | | | | | | Mean of AUC of each model on all 10 sites | SD of AUC of each model on all 10 sites |
|---|---|---|---|---|---|---|---|---|---|---|---|---|---|---|---|---|---|---|---|---|---|---|
| | | SGH1 | | SGH2 | | SGH3 | | SGH4 | | SGH5 | | SGH6 | | SGH7 | | SGH8 | | SGH9 | | SGH10 | | | |
| | | AUC | CI | AUC | CI | AUC | CI | AUC | CI | AUC | CI | AUC | CI | AUC | CI | AUC | CI | AUC | CI | AUC | CI | | |
| Model 1[a] | 7 | 0.7730 | 0.6936-0.8523 | 0.7387 | 0.6659-0.8114 | 0.7473 | 0.6839-0.8106 | 0.7124 | 0.6557-0.7690 | 0.7023 | 0.6479-0.7567 | 0.7618 | 0.7113-0.8123 | 0.7329 | 0.6874-0.7784 | 0.7491 | 0.7018-0.7964 | 0.7651 | 0.7261-0.8040 | 0.7526 | 0.7139-0.7913 | 0.7435 | 0.0226 |
| Model 2[b] | 5 | 0.7628 | 0.6841-0.8416 | 0.7427 | 0.6708-0.8147 | 0.7556 | 0.6950-0.8162 | 0.7087 | 0.6497-0.7677 | 0.7239 | 0.6645-0.7833 | 0.7645 | 0.7128-0.8163 | 0.7305 | 0.6814-0.7797 | 0.7418 | 0.6952-0.7885 | 0.7402 | 0.6976-0.7828 | 0.7289 | 0.6853-0.7726 | 0.7400 | 0.0177 |
| Model 3[c] | 7 | 0.7745 | 0.6954-0.8537 | 0.7153 | 0.6386-0.7920 | 0.7915 | 0.7353-0.8477 | 0.7277 | 0.6713-0.7842 | 0.7479 | 0.6941-0.8017 | 0.7723 | 0.7217-0.8229 | 0.7616 | 0.7160-0.8071 | 0.7579 | 0.7098-0.8060 | 0.7604 | 0.7189-0.8020 | 0.7683 | 0.7271-0.8095 | 0.7577 | 0.0225 |
| Model 4[d] | 7 | 0.7967 | 0.7243-0.8690 | 0.7427 | 0.6667-0.8187 | 0.7752 | 0.7145-0.8358 | 0.7211 | 0.6647-0.7774 | 0.7293 | 0.6742-0.7844 | 0.7579 | 0.7008-0.8149 | 0.7367 | 0.6893-0.7841 | 0.7400 | 0.6878-0.7923 | 0.7530 | 0.7099-0.7961 | 0.7712 | 0.7334-0.8091 | 0.7524 | 0.0233 |
| Model 5[e] | 4 | 0.7512 | 0.6659-0.8366 | 0.7401 | 0.6688-0.8114 | 0.7527 | 0.6908-0.8146 | 0.6917 | 0.6334-0.7500 | 0.7328 | 0.6765-0.7891 | 0.7492 | 0.6948-0.8035 | 0.7348 | 0.6889-0.7808 | 0.7223 | 0.6721-0.7724 | 0.7519 | 0.7105-0.7933 | 0.7547 | 0.7139-0.7956 | 0.7381 | 0.0195 |
| Model 6[f] | 7 | 0.7787 | 0.6969-0.8605 | 0.7406 | 0.6643-0.8170 | 0.7714 | 0.7111-0.8317 | 0.7209 | 0.6633-0.7786 | 0.7241 | 0.6685-0.7797 | 0.7710 | 0.7181-0.8240 | 0.7509 | 0.7055-0.7963 | 0.7528 | 0.7033-0.8023 | 0.7769 | 0.7357-0.8180 | 0.7979 | 0.762-0.8338 | 0.7585 | 0.02508 |
| Model 7[g] | 8 | 0.7795 | 0.7019-0.8571 | 0.7352 | 0.6576-0.8127 | 0.7755 | 0.7147-0.8364 | 0.7204 | 0.6608-0.7801 | 0.7254 | 0.6666-0.7842 | 0.7804 | 0.7258-0.8350 | 0.7300 | 0.6833-0.7767 | 0.7453 | 0.6967-0.7939 | 0.7696 | 0.7288-0.8104 | 0.7810 | 0.7429-0.8190 | 0.7542 | 0.0252 |
| Model 8[h] | 8 | 0.7666 | 0.6826-0.8507 | 0.7317 | 0.6546-0.8087 | 0.7669 | 0.7049-0.8288 | 0.7161 | 0.6592-0.7731 | 0.7189 | 0.6610-0.7768 | 0.7895 | 0.7404-0.8386 | 0.7444 | 0.6995-0.7892 | 0.7498 | 0.7005-0.7992 | 0.7762 | 0.7378-0.8145 | 0.7818 | 0.7434-0.8202 | 0.7542 | 0.0261 |
| Model 9[i] | 6 | 0.7868 | 0.7126-0.8610 | 0.7430 | 0.6689-0.8171 | 0.7613 | 0.6987-0.8239 | 0.7033 | 0.6426-0.7640 | 0.7330 | 0.6788-0.7872 | 0.7628 | 0.7078-0.8177 | 0.7629 | 0.7181-0.8077 | 0.7608 | 0.7145-0.8071 | 0.7674 | 0.7273-0.8074 | 0.7759 | 0.7364-0.8155 | 0.7557 | 0.0238 |
| Model 10[j] | 7 | 0.7965 | 0.7240-0.8689 | 0.7442 | 0.6679-0.8204 | 0.7783 | 0.7198-0.8367 | 0.7148 | 0.6591-0.7706 | 0.7348 | 0.6807-0.7888 | 0.7734 | 0.7210-0.8258 | 0.7478 | 0.7013-0.7944 | 0.7674 | 0.7215-0.8132 | 0.7806 | 0.7412-0.8199 | 0.7889 | 0.7512-0.8267 | 0.7627 | 0.0262 |
| Model Federated[k] | 8 | 0.7834 | 0.7085-0.8583 | 0.7372 | 0.6621-0.8124 | 0.7604 | 0.6992-0.8217 | 0.7285 | 0.6730-0.7839 | 0.7433 | 0.6908-0.7958 | 0.7758 | 0.7239-0.8278 | 0.7804 | 0.7381-0.8228 | 0.7640 | 0.7178-0.8101 | 0.7749 | 0.7338-0.8160 | 0.7850 | 0.7471-0.8229 | 0.7633 | 0.0204 |
| Model Pooled[l] | 8 | 0.7405 | 0.6603-0.8206 | 0.7303 | 0.6520-0.8085 | 0.7665 | 0.7004-0.8326 | 0.7195 | 0.6621-0.7768 | 0.7499 | 0.6975-0.8023 | 0.7992 | 0.7459-0.8525 | 0.7835 | 0.7307-0.8363 | 0.7745 | 0.7300-0.8191 | 0.7601 | 0.7208-0.7994 | 0.8074 | 0.7658-0.8490 | 0.7631 | 0.0289 |
| Average AUC of all 10 local models on each site | | 0.77663 | | 0.73742 | | 0.76757 | | 0.71371 | | 0.72724 | | 0.76828 | | 0.74325 | | 0.74872 | | 0.76413 | | 0.77012 | | | |

[a-j] Local model obtained via AutoScore independently on site 1 to 10
[k] Federated model obtained via FedScore
[l] Pooled model obtained via AutoScore

**eTable 2:** Scoring tables
(a) Scoring table of local model generated on site 1 via AutoScore.

| Variable | Interval | Point |
| --- | --- | --- |
| Pulse rate (per minute) | <70 | 0 |
|  | [70,100) | 1 |
|  | [100,120) | 3 |
|  | >=120 | 6 |
| Systolic blood pressure (mmHg) | <98 | 6 |
|  | [98,114) | 8 |
|  | [114,160) | 4 |
|  | [160,188) | 0 |
|  | >=188 | 3 |
| Age (years) | <30 | 0 |
|  | [30,50) | 61 |
|  | [50,79) | 69 |
|  | [79,88) | 71 |
|  | >=88 | 72 |
| Diastolic blood pressure (mmHg) | <52 | 3 |
|  | [52,60) | 1 |
|  | [60,83) | 0 |
|  | [83,96) | 3 |
|  | >=96 | 1 |
| Oxygen saturation (%) | <94 | 4 |
|  | [94,96) | 1 |
|  | >=96 | 0 |
| Respiration rate (per minute) | <16 | 3 |
|  | [16,17) | 0 |
|  | [17,19) | 2 |
|  | [19,22) | 4 |
|  | >=22 | 2 |
| Number of emergency admissions in the past year | <2 | 0 |
|  | [2,5) | 2 |
|  | >=5 | 4 |

(b) Scoring table of local model generated on site 2 via AutoScore.

| Variable | Interval | Point |
| --- | --- | --- |
| Pulse rate (per minute) | <60 | 1 |
|  | [60,71) | 0 |
|  | [71,100) | 1 |
|  | >=100 | 5 |
| Systolic blood pressure (mmHg) | <99 | 12 |
|  | [99,114) | 9 |
|  | [114,158) | 8 |
|  | >=158 | 0 |
| Age (years) | <29 | 0 |
|  | [29,49) | 63 |
|  | [49,79) | 71 |
|  | [79,88) | 73 |
|  | >=88 | 75 |
| Diastolic blood pressure (mmHg) | <53 | 2 |
|  | [53,60) | 1 |
|  | [60,83) | 0 |
|  | [83,96) | 2 |
|  | >=96 | 1 |
| Oxygen saturation (%) | <94 | 5 |
|  | [94,96) | 1 |
|  | [96,99) | 0 |
|  | >=99 | 1 |

(c) Scoring table of local model generated on site 3 via AutoScore.

| Variable | Interval | Point |
|---|---|---|
| Pulse rate (per minute) | <60 | 1 |
|  | [60,71) | 0 |
|  | [71,99) | 1 |
|  | [99,117) | 8 |
|  | >=117 | 15 |
| Systolic blood pressure (mmHg) | <98 | 19 |
|  | [98,114) | 13 |
|  | [114,159) | 9 |
|  | [159,188) | 0 |
|  | >=188 | 4 |
| Age (years) | <29 | 0 |
|  | [29,48) | 10 |
|  | [48,79) | 19 |
|  | [79,88) | 25 |
|  | >=88 | 30 |
| Diastolic blood pressure (mmHg) | <53 | 11 |
|  | [53,61) | 10 |
|  | [61,96) | 6 |
|  | >=96 | 0 |
| Oxygen saturation (%) | <93 | 10 |
|  | [93,99) | 0 |
|  | >=99 | 1 |
| Respiration rate (per minute) | <19 | 0 |
|  | [19,22) | 3 |
|  | >=22 | 7 |
| Number of emergency admissions in the past year | <2 | 0 |
|  | [2,4) | 8 |
|  | >=4 | 4 |

(d) Scoring table of local model generated on site 4 via AutoScore.

| Variable | Interval | Point |
|---|---|---|
| Pulse rate (per minute) | <60 | 1 |
| | [60,71) | 0 |
| | [71,100) | 1 |
| | [100,119) | 4 |
| | >=119 | 5 |
| Systolic blood pressure (mmHg) | <98 | 13 |
| | [98,115) | 10 |
| | [115,160) | 9 |
| | [160,187) | 5 |
| | >=187 | 0 |
| Age (years) | <28 | 0 |
| | [28,50) | 60 |
| | [50,80) | 64 |
| | [80,89) | 66 |
| | >=89 | 68 |
| Diastolic blood pressure (mmHg) | <52 | 3 |
| | [52,61) | 1 |
| | [61,84) | 0 |
| | [84,97) | 1 |
| | >=97 | 3 |
| Oxygen saturation (%) | <94 | 4 |
| | [94,96) | 0 |
| | [96,99) | 1 |
| | >=99 | 2 |
| Respiration rate (per minute) | <16 | 4 |
| | [16,17) | 0 |
| | [17,19) | 2 |
| | [19,22) | 3 |
| | >=22 | 5 |
| Number of emergency admissions in the past year | <2 | 0 |
| | [2,5) | 3 |
| | >=5 | 2 |

(e) Scoring table of local model generated on site 5 via AutoScore.

| Variable | Interval | Point |
|---|---|---|
| Pulse rate (per minute) | <60 | 1 |
| | [60,71) | 0 |
| | [71,100) | 5 |
| | [100,119) | 17 |
| | >=119 | 24 |
| Systolic blood pressure (mmHg) | <98 | 24 |
| | [98,114) | 18 |
| | [114,160) | 6 |
| | [160,188) | 0 |

| | >=188 | 3 |
|---|---|---|
| Age (years) | <29 | 3 |
| | [29,49) | 0 |
| | [49,79) | 30 |
| | [79,88) | 41 |
| | >=88 | 43 |
| Diastolic blood pressure (mmHg) | <52 | 9 |
| | [52,61) | 2 |
| | [61,84) | 1 |
| | >=84 | 0 |

(f) Scoring table of local model generated on site 6 via AutoScore.

| Variable | Interval | Point |
|---|---|---|
| Pulse rate (per minute) | <71 | 0 |
| | [71,100) | 2 |
| | [700,120) | 9 |
| | >=120 | 14 |
| Systolic blood pressure (mmHg) | <98 | 16 |
| | [98,114) | 15 |
| | [114,159) | 7 |
| | [159,186) | 6 |
| | >=186 | 0 |
| Age (years) | <28 | 0 |
| | [28,48) | 13 |
| | [48,80) | 28 |
| | [80,88) | 30 |
| | >=88 | 32 |
| Diastolic blood pressure (mmHg) | <52 | 11 |
| | [52,84) | 4 |
| | [84,97) | 0 |
| | >=97 | 6 |
| Oxygen saturation (%) | <94 | 9 |
| | [94,99) | 4 |
| | >=99 | 0 |
| Respiration rate (per minute) | <16 | 9 |
| | [16,17) | 0 |
| | [17,19) | 2 |
| | [19,22) | 5 |
| | >=22 | 11 |
| Number of emergency admissions in the past year | <2 | 0 |
| | [2,4) | 6 |
| | >=4 | 3 |

(g) Scoring table of local model generated on site 7 via AutoScore.

| Variable | Interval | Point |
|---|---|---|

| Variable | Interval | Point |
| --- | --- | --- |
| Pulse rate (per minute) | <60 | 5 |
| | [60,71) | 0 |
| | [71,100) | 2 |
| | >=100 | 9 |
| Systolic blood pressure (mmHg) | <98 | 31 |
| | [98,114) | 24 |
| | [114,160) | 16 |
| | [160,189) | 7 |
| | >=189 | 0 |
| Age (years) | <30 | 0 |
| | [30,49) | 7 |
| | [49,79) | 18 |
| | [79,88) | 22 |
| | >=88 | 25 |
| Diastolic blood pressure (mmHg) | <52 | 4 |
| | [52,61) | 3 |
| | [61,84) | 0 |
| | [84,97) | 3 |
| | >=97 | 7 |
| Oxygen saturation (%) | <94 | 7 |
| | [94,96) | 3 |
| | [94,99) | 0 |
| | >=99 | 3 |
| Respiration rate (per minute) | <16 | 12 |
| | [16,17) | 0 |
| | [17,19) | 5 |
| | [19,22) | 8 |
| | >=22 | 15 |
| Number of emergency admissions in the past year | <2 | 0 |
| | [2,5) | 5 |
| | >=5 | 4 |
| Day of week | Friday | 1 |
| | Midweek | 0 |
| | Other | 2 |

(h) Scoring table of local model generated on site 8 via AutoScore.

| Variable | Interval | Point |
| --- | --- | --- |
| Pulse rate (per minute) | <61 | 0 |
| | [61,71) | 2 |
| | [71,100) | 0 |
| | [100,118) | 5 |
| | >=118 | 10 |
| Systolic blood pressure (mmHg) | <98 | 15 |
| | [98,114) | 11 |
| | [114,160) | 3 |
| | [160,189) | 2 |

| Variable | Interval | Point |
|---|---|---|
| | >=189 | 0 |
| Age (years) | <28 | 0 |
| | [28,49) | 18 |
| | [48,79) | 31 |
| | [79,88) | 35 |
| | >=88 | 39 |
| Diastolic blood pressure (mmHg) | <52 | 5 |
| | [52,61) | 3 |
| | [61,84) | 1 |
| | [84,97) | 0 |
| | >=97 | 4 |
| Oxygen saturation (%) | <96 | 7 |
| | [96,99) | 0 |
| | >=99 | 2 |
| Respiration rate (per minute) | <16 | 8 |
| | [16,17) | 0 |
| | [17,19) | 3 |
| | [19,22) | 5 |
| | >=22 | 13 |
| Number of emergency admissions in the past year | <2 | 0 |
| | [2,5) | 9 |
| | >=5 | 2 |
| Day of week | Midweek | 0 |
| | Monday | 3 |
| | Other | 2 |

(i) Scoring table of local model generated on site 9 via AutoScore.

| Variable | Interval | Point |
|---|---|---|
| Pulse rate (per minute) | <60 | 1 |
| | [60,71) | 0 |
| | [71,100) | 4 |
| | [100,119) | 10 |
| | >=119 | 16 |
| Systolic blood pressure (mmHg) | <98 | 21 |
| | [98,115) | 13 |
| | [115,159) | 8 |
| | [159,187) | 5 |
| | >=187 | 0 |
| Age (years) | <29 | 0 |
| | [29,49) | 14 |
| | [49,79) | 31 |
| | [79,88) | 36 |
| | >=88 | 39 |
| Diastolic blood pressure (mmHg) | <52 | 5 |
| | [52,61) | 2 |
| | [61,84) | 0 |

| | [84,96) | 2 |
| | >=96 | 3 |
| Oxygen saturation (%) | <94 | 11 |
| | [94,96) | 1 |
| | [96,99) | 0 |
| | >=99 | 1 |
| Respiration rate (per minute) | <16 | 0 |
| | [16,17) | 3 |
| | [17,19) | 2 |
| | [19,22) | 5 |
| | >=22 | 8 |

(j) Scoring table of local model generated on site 10 via AutoScore.

| **Variable** | **Interval** | **Point** |
| --- | --- | --- |
| Pulse rate (per minute) | <60 | 3 |
| | [60,71) | 0 |
| | [71,100) | 2 |
| | [100,119) | 8 |
| | >=119 | 14 |
| Systolic blood pressure (mmHg) | <99 | 22 |
| | [99,114) | 15 |
| | [114,158) | 10 |
| | [158,186) | 6 |
| | >=186 | 0 |
| Age (years) | <29 | 0 |
| | [29,49) | 13 |
| | [49,79) | 25 |
| | [79,88) | 29 |
| | >=88 | 30 |
| Diastolic blood pressure (mmHg) | <53 | 5 |
| | [53,83) | 3 |
| | [83,96) | 0 |
| | >=96 | 3 |
| Oxygen saturation (%) | <94 | 9 |
| | [94,96) | 1 |
| | >=96 | 0 |
| Respiration rate (per minute) | <16 | 3 |
| | [16,17) | 0 |
| | [17,19) | 4 |
| | [19,22) | 8 |
| | >=22 | 13 |
| Number of emergency admissions in the past year | <2 | 0 |
| | [2,4) | 7 |
| | >=4 | 4 |

(k) Scoring table of federated model generated via FedScore.

| Variable | Interval | Point |
| --- | --- | --- |
| Pulse rate (per minute) | <70 | 13 |
| | [70,100) | 15 |
| | [100,118) | 19 |
| | >=118 | 21 |
| Systolic blood pressure (mmHg) | <98 | 13 |
| | [98,114) | 8 |
| | [114,159) | 5 |
| | [159,187) | 2 |
| | >=187 | 0 |
| Age (years) | <28 | 13 |
| | [28,48) | 22 |
| | [48,79) | 31 |
| | [79,88) | 34 |
| | >=88 | 37 |
| Diastolic blood pressure (mmHg) | <52 | 13 |
| | [52,60) | 11 |
| | [60,83) | 10 |
| | >=83 | 12 |
| Oxygen saturation (%) | <93 | 13 |
| | [93,96) | 8 |
| | >=96 | 6 |
| Respiration rate (per minute) | <16 | 13 |
| | [16,17) | 16 |
| | [17,19) | 14 |
| | [19,22) | 16 |
| | >=22 | 18 |
| Number of emergency admissions in the past year | <1 | 13 |
| | [1,4) | 16 |
| | >=4 | 17 |
| Day of Week | Friday | 13 |
| | Other | 12 |

(l) Pooled model obtained via AutoScore.

| Variable | Interval | Point |
| --- | --- | --- |
| Pulse rate (per minute) | <60 | 2 |
| | [60,71) | 0 |
| | [71,100) | 2 |
| | [100,119) | 9 |
| | >=119 | 13 |
| Systolic blood pressure (mmHg) | <98 | 19 |
| | [98,114) | 14 |
| | [114,159) | 7 |
| | [159,188) | 1 |
| | >=188 | 0 |

| Age (years) | <29 | 0 |
| | [29,49) | 16 |
| | [49,79) | 30 |
| | [79,88) | 34 |
| | >=88 | 37 |
| Diastolic blood pressure (mmHg) | <52 | 4 |
| | [52,61) | 1 |
| | [61,84) | 0 |
| | [84,97) | 1 |
| | >=97 | 2 |
| Oxygen saturation (%) | <94 | 9 |
| | [94,96) | 2 |
| | >=96 | 0 |
| Respiration rate (per minute) | <16 | 8 |
| | [16,17) | 0 |
| | [17,19) | 2 |
| | [19,22) | 5 |
| | >=22 | 11 |
| Number of emergency admissions in the past year | <2 | 0 |
| | [2,5) | 6 |
| | >=5 | 4 |
| Day of Week | Friday | 0 |
| | Other | 1 |

# eFigure 1

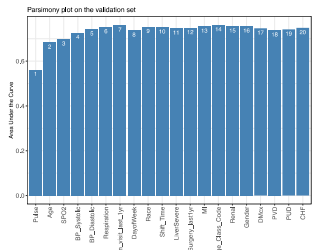
(a)

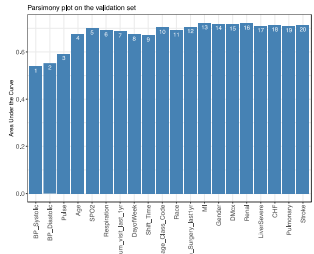
(b)

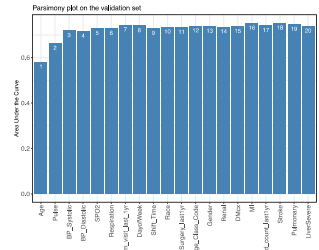
(c)

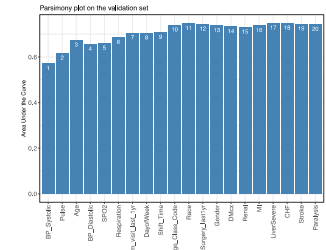
(d)

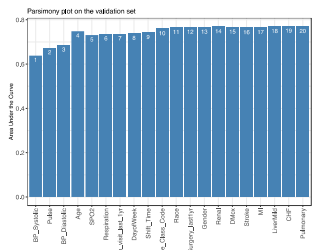
(e)

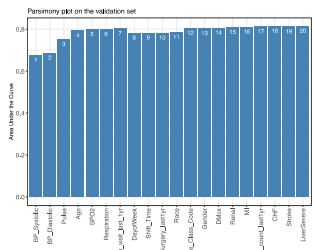
(f)

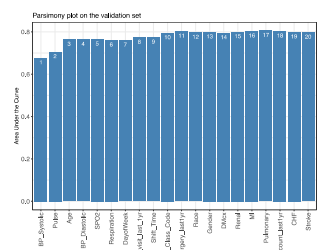
(g)

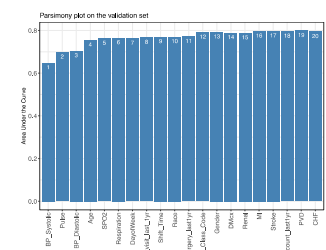
(h)

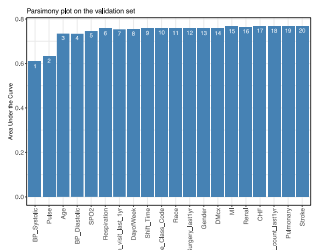
(i)

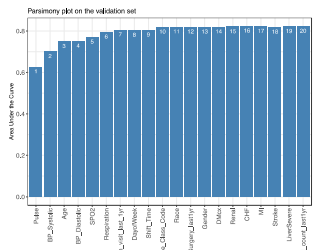
(j)

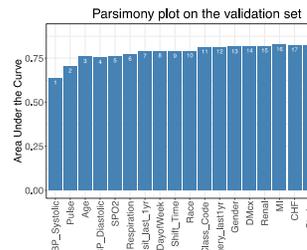
(k)

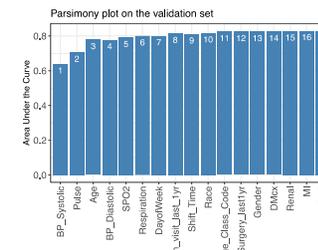
(l)

Abbreviations:
AUC, area under the curve; CI, confidence interval; SD, standard deviation; SBP, systolic blood pressure; DBP, diastolic blood pressure; SpO$_2$, oxygen saturation as measured by pulse oximetry; PACS, Patient Acuity Category Scale; ED, emergency department